\documentclass[letterpaper]{article} 
\usepackage{aaai2026}  
\usepackage{times}  
\usepackage{helvet}  
\usepackage{courier}  
\usepackage[hyphens]{url}  
\usepackage{graphicx} 
\usepackage{natbib}  
\usepackage{caption} 
\usepackage{subcaption} 
\usepackage{algorithm}
\usepackage{algorithmic}

\usepackage{booktabs}
\usepackage{multirow}
\usepackage{array}
\usepackage{amsmath}
\usepackage{amsthm}

\newtheorem{theorem}{Theorem}

\usepackage{tikz}

\usepackage{newfloat}
\usepackage{listings}
\DeclareCaptionStyle{ruled}{labelfont=normalfont,labelsep=colon,strut=off} 
\floatstyle{ruled}
\newfloat{listing}{tb}{lst}{}
\floatname{listing}{Listing}
\title{Bridging Cognitive Gap: Hierarchical Description Learning for Artistic Image Aesthetics Assessment}
\author {
    Henglin Liu\textsuperscript{\rm 1, 2}\equalcontrib, Nisha Huang\textsuperscript{\rm 1, 3}\equalcontrib, Chang Liu\textsuperscript{\rm 1}\footnotemark[2], Jiangpeng Yan\textsuperscript{\rm 1, 5}, Huijuan Huang\textsuperscript{\rm 2}, Jixuan Ying\textsuperscript{\rm 1}, Tong-Yee Lee\textsuperscript{\rm 4}, Pengfei Wan\textsuperscript{\rm 2}, Xiangyang Ji\textsuperscript{\rm 1}\thanks{Corresponding author.}
}
\affiliations {
    \textsuperscript{\rm 1}Tsinghua University \\
    \textsuperscript{\rm 2}Kling Team, Kuaishou Technology \\
    \textsuperscript{\rm 3}Pengcheng Laboratory \\
    \textsuperscript{\rm 4}National Cheng Kung University \\
    \textsuperscript{\rm 5}E Fund Management Co., Ltd. \\
    \{liu-hl24,hns24,yingjx23\}@mails.tsinghua.edu.cn, yanjp13@tsinghua.org.cn, 
    \{liuchang2022,xyji\}@tsinghua.edu.cn, huanghuijuan.thu@gmail.com, tonylee@ncku.edu.tw, wanpengfei@kuaishou.com
}

\begin{document}

\maketitle

\begin{abstract}
The aesthetic quality assessment task is crucial for developing a human-aligned quantitative evaluation system for AIGC. However, its inherently complex nature—spanning visual perception, cognition, and emotion—poses fundamental challenges. Although aesthetic descriptions offer a viable representation of this complexity, two critical challenges persist:  (1) data scarcity and imbalance: existing dataset overly focuses on visual perception and neglects deeper dimensions due to the expensive manual annotation; and (2) model fragmentation: current visual networks isolate aesthetic attributes with multi-branch encoder, while multimodal methods represented by contrastive learning struggle to effectively process long-form textual descriptions. To resolve challenge (1), we first present the Refined Aesthetic Description (RAD) dataset, a large-scale (70k), multi-dimensional structured dataset, generated via an iterative pipeline without heavy annotation costs and easy to scale. To address challenge (2), we propose ArtQuant, an aesthetics assessment framework for artistic image which not only couple isolated aesthetic dimensions through joint description generation, but also better model long-text semantics with the help of LLM decoders. Besides, theoretical analysis confirms this symbiosis: RAD's semantic adequacy (data) and generation paradigm (model) collectively minimize prediction entropy, providing mathematical grounding for the framework. Our approach achieves state-of-the-art performance on several datasets while requiring only 33\% of conventional training epochs, narrowing the cognitive gap between artistic image and aesthetic judgment. The code is available at: \url{https://github.com/Henglin-Liu/ArtQuant}.
\end{abstract}

\section{Introduction}
\begin{quotation}
\emph{``Painting is a mental thing, a thought.''
\begin{flushright}
--- Leonardo Da Vinci
\end{flushright}
}
\end{quotation}
With the rapid development of computer vision, remarkable progress has been made in image generation and personalization, making artistic creation~\cite{huang2025creativesynth,huang2025artcrafter} more accessible to the public. However, the development of quantitative evaluation of artwork aesthetics relatively lags.
Among these, aesthetic quality assessment plays a pivotal role, as it bridges the gap between machine-generated outputs and human judgments. 
\begin{figure}
  \centering
  \includegraphics[width=\linewidth]{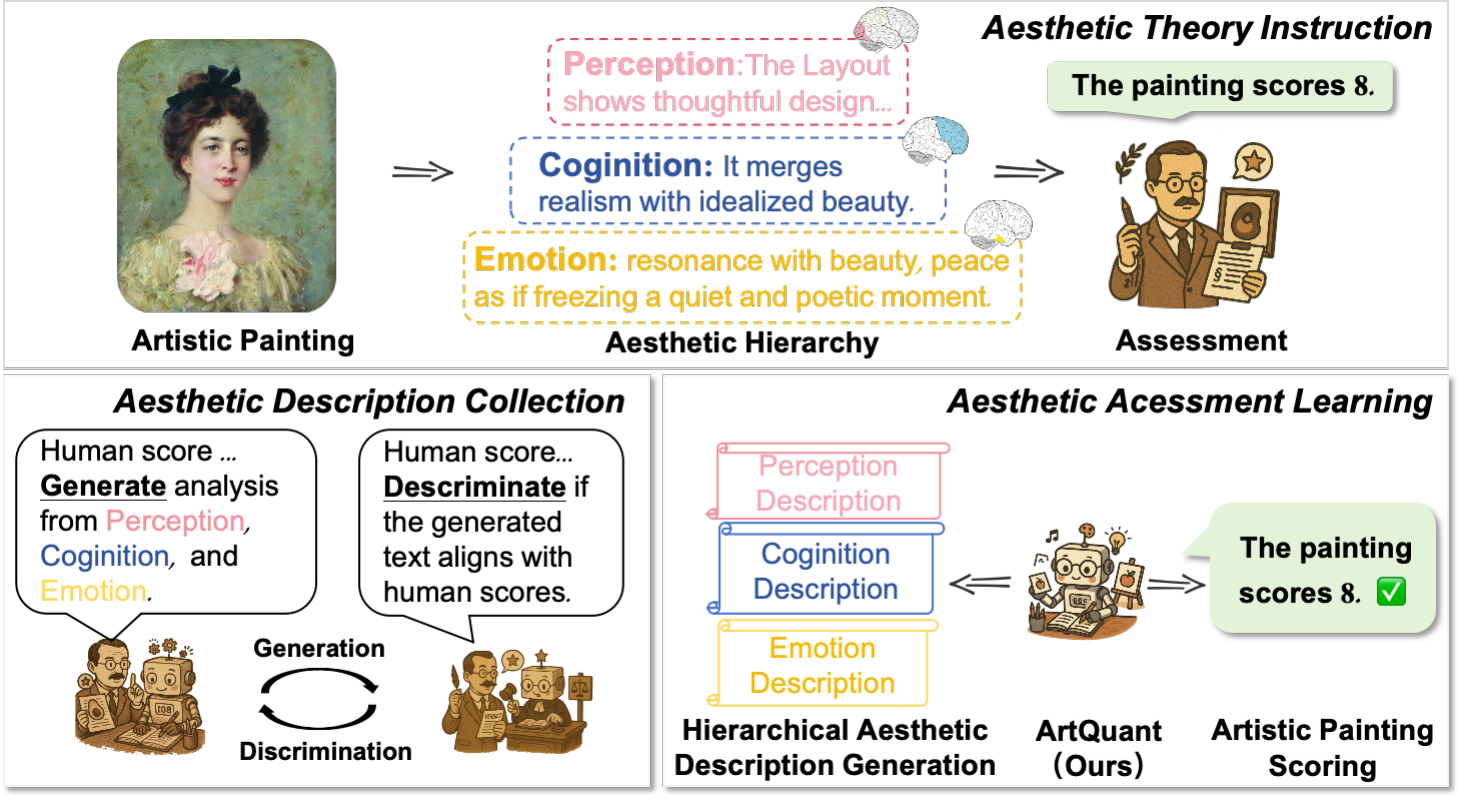}
  \caption{Top: Hierarchical cognitive mechanism of human artistic aesthetics, serving as our theoretical foundation. Bottom left: we employ an iterative data collection framework to optimize the training data distribution, ensuring that the generated aesthetic analysis aligns with human ratings. Bottom right: Dual-task learning (description generation + scoring prediction) enhances model's alignment with human judgement.}
  \label{limitation}
\end{figure}

This task is fundamentally challenging due to its inherently multidimensional nature. 
Prior work has typically addressed it through specialized architectures: Amirshahi et al.~\cite{Amirshahi2017JudgingAQ} developed color-based features with an SVM classifier for artistic quality assessment. Chen et al.~\cite{10810501} proposed AKA-Net, employing multi-branch networks to separately model color, composition, and content dimensions, while PVAFE~\cite{tang2025artistic} decomposed aesthetics into content, vivid color, and color harmony components. Similarly, ArtCLIP~\cite{jin2024apddv2} introduced an assessment framework using contrastive learning on image-comment pairs, but required separate training for each of 10 aesthetic attributes.

However, current AIAA (Artistic Image Aesthetics Assessment) methods are not well-aligned with the human aesthetic process. Studies~\cite{aesthetic,leder2004model,chatterjee2003prospects,huang2024diffstyler,huang2025mate} indicate that human aesthetic experience is the result of multiple processes, including perceptual, cognitive, and emotional processes. Hierarchical aesthetic descriptions effectively capture the progression, combining comprehensive feature with analytical flexibility to provide a promising solution to address the existing misalignment. However, this direction remains unexplored in prior work, due to substantial challenges in both data and methodological design. (1) \textbf{Data}: the collection of human-annotated aesthetic descriptions remains prohibitively expensive, and existing dataset, APDD~\cite{zhang2024long}, is overly generalized with insufficient visual evidence and limited to perception technical analysis while neglecting deeper aesthetic dimensions, which is essential for artistic image aesthetics assessment (as shown in Fig.~\ref{limitation}). (2) \textbf{Method}: the existing method, such as contrastive learning architectures~\cite{jin2024apddv2} struggle to effectively process long-form textual descriptions. 
This raises three core research questions. \textbf{(1) Can hierarchical descriptive texts serve as effective carriers of aesthetic prior knowledge? (2) What are the specific contributions of different levels of textual information to performance improvement? (3) Beyond the sufficiency of textual information itself, are there other mechanisms that regulate the aesthetic capabilities of MLLM?} An in-depth exploration of these questions will contribute to a deeper understanding of the role of textual information for AIAA.

To explore these questions, we propose ArtQuant, the first MLLM tailored for the AIAA task, specializing in aesthetic score prediction. We first present an aesthetic description generation pipeline to alleviate the scarcity of aesthetic description data. We propose a multi-level description generation framework inspired by the data annotation approach~\cite{tan2024large}. Specifically, we get distribution on certain dataset for better human-preference-aligned and create an aesthetic template for systematic and comprehensive generation. To ensure the quality of the generated content, we adopt the alignment score with human aesthetic ratings judged by LLM as an evaluation metric~\cite{li2024generation}. The framework not only preserves the statistical patterns of the original data but also enhances the rationality and diversity of the generated comments. Using this framework, we constructed the \textbf{R}efined \textbf{A}esthetic \textbf{D}escriptions (RAD) dataset, which contains a total of 70k comments data.

Building upon the RAD, we introduce a novel auxiliary description task to effectively learn aesthetic representations. This is based on a key hypothesis: generating aesthetic descriptions enables more comprehensive understanding of artistic elements. By training the model to produce textual descriptions of art images, it is forced to explicitly decompose aesthetic elements such as brushstroke, lighting, and mood. This decomposition yields interpretable features that serve as aesthetic prior knowledge for score prediction. Notably, more complete description levels lead to better score prediction performance. 

\begin{figure}[]
  \centering
  \includegraphics[width=\linewidth]{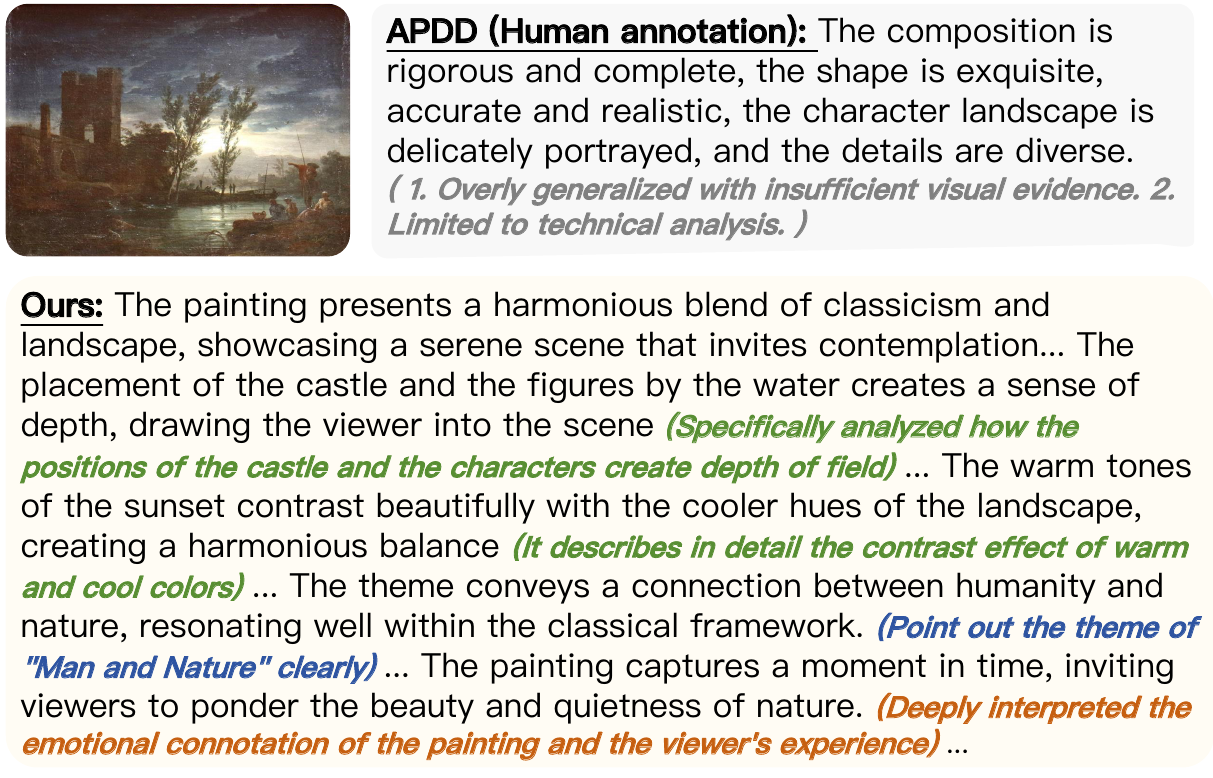}
  \caption{Grey font indicates limitations in the existing dataset APDD~\cite{jin2024apddv2}. The green, blue, and orange fonts represent Perception, Cognition and Emotion respectively, demonstrating the fine-grained and multi-level nature of our methodology.}
  \label{limitation}
\end{figure}

We formalize this intuition via information theory, proving that learning auxiliary descriptions inherently bounds scoring error. Specifically, the prediction uncertainty \( H(Y|Z) \) is constrained by two key factors: (1) description sufficiency (\( H(Y|D) \)): the relevance of generated text to scoring, guaranteed by our dataset’s curated descriptions and (2) representation quality (\( H(D|Z) \)): the model’s ability to reconstruct accurate descriptions from images, ensured by our pretraining objective.

Besides, a main challenge is predicting continuous scores with MLLMs, which are built for discrete token outputs. Previous approaches~\cite{wu2023q,you2025teaching} struggle to fit real labels accurately and handle datasets lacking variance information (common in art datasets like APDD~\cite{jin2024apddv2} and BAID~\cite{yi2023towards}). Inspired by mapping, we propose Score-Based Distribution Estimation method. It models numerical outputs as expected values of discrete token probabilities by minimizing the error between the target distribution's expectation and the ground truth. This approach inherently reduce label errors while eliminate the dependence on variance label.

To validate the effectiveness of our method, we test it on three common AIAA datasets. Beyond achieving state-of-the-art performance, our model converges with fewer training epochs. Our main contributions are as follows:

\begin{itemize}
    \item To overcome the rigid and isolated nature of traditional aesthetic modeling approaches, we propose a dual-level solution: (1) At the data level, we develop a scalable framework featuring multi-level generation with human-preference-aligned aesthetic templates and LLM-validated quality control (§3.1); (2) At the methodology level, we introduce an auxiliary description generation task that explicitly decomposes and learns fine-grained aesthetic features as interpretable intermediate representations for scoring and the Score-Based Distribution Estimation method to better model scoring (§3.2).
    \item We formalize this intuition with information-theoretic proof that auxiliary description learning bounds scoring error via description sufficiency (\( H(Y|D) \)) and representation quality (\( H(D|Z) \)) (§4).
    \item Extensive experiments demonstrate our method's superiority over baseline approaches on three datasets, within just 33\% of the training epochs required by conventional approaches (§5). 

\end{itemize}

\section{Related Work}
\subsection{Artistic Image Aesthetics Assessment}
Recent advances in artistic image aesthetics assessment (AIAA) have evolved from manual feature extraction to data-driven approaches.  
Initial work relied on manually designed features to quantify aesthetic properties. Li et al.~\cite{li2009aesthetic} proposed a segmentation-based model to capture global composition and local attractiveness in paintings separately. 
The advent of large-scale datasets and self-supervision marked a significant shift. Yi et al.~\cite{yi2023towards} contributed to the BAID dataset (60,000+ artworks) and the SAAN architecture, which jointly learns style and aesthetic features without explicit labels. Further refinements incorporated multi-attribute supervision: Chen et al.~\cite{10810501} proposed AKA-Net to amalgamate attribute-specific knowledge. Jin et al.~\cite{jin2024apddv2} introduce ArtCLIP, a style-specific art assessment framework that employs contrastive learning on paired artistic images and short aesthetic comments. While recent work like ArtCLIP uses aesthetic comments, existing methods underutilize rich, semantically meaningful descriptions. To bridge this gap, we propose auxiliary description learning for multimodal foundation models, which leverages fine-grained textual annotations to enhance aesthetic understanding.
\subsection{Multimodal Large Language Models}
Recent advances in Multimodal Large Language Models (MLLMs)~\cite{liu2023visual,bai2023qwen,bai2025qwen2} have demonstrated remarkable capabilities in bridging vision and language modalities. In the domain of artistic image assessment, systems like AesExpert~\cite{huang2024aesexpert} leverage rich aesthetic critique databases to fine-tune multimodal foundation models, enabling aesthetic-related question answering. Similarly, GalleryGPT~\cite{bin2024gallerygpt} harnesses large multimodal models' perceptual and generative strengths to produce comprehensive art analyses. However, none of these existing approaches provide quantifiable aesthetic scoring, significantly restricting their practical utility in real-world applications. 


In contrast to the aforementioned works, our approach leverages large language models to generate multi-level aesthetic descriptions, better capturing the hierarchical nature of human aesthetic perception and significantly enhancing quantitative scoring capabilities for artworks.

\section{Method}
Our approach comprises two core parts: (1) Aesthetic description collection and (2) Aesthetic Acessment Learning.
\subsection{Aesthetic Description Collection}

\label{rad_dataset}
To produce meaningful aesthetic descriptions for score training, our hierarchical generation pipeline comprises three key components: (1) aesthetic data preprocessing that eliminates systematic bias using dataset statistics and emulates human judgment hierarchy, (2) structured description generation that synthesizes multi-level aesthetic analysis, and (3) discriminative quality control that ensures score-description consistency through iterative refinement.

\noindent\textbf{Aesthetic data preprocessing.}
Instead of directly using artistic images as visual inputs and relying solely on artistic scores as human aesthetic guidance, we introduce two additional preprocessing steps. Due to differences in the annotating population, different datasets often have different data distributions. Samples with high absolute scores but low relative scores can mislead large models to generate descriptions with wrong preferences. Therefore, we need to combine the scores with the dataset distribution for correction. We incorporate dataset-level statistics (mean, median, and variance) as additional conditioning inputs to the MLLM. To emulate the hierarchical nature of human artistic aesthetic judgment~\cite{aesthetic}, we propose a three-level generative framework that systematically progresses from perception to cognitive processing and emotional evaluation, mirroring the staged refinement of aesthetic appreciation in human experience. 

\noindent\textbf{Description generation.}
After preprocessing, we obtain a quadruple $(x_v, s, D, T)$, where $x_v$ denotes the visual output, $s \in [0, M]$ its quality score, $D \sim \mathcal{N}(\mu, \sigma^2)$ the dataset's score distribution, and $T$ a set of aesthetic templates. These inputs are processed by $\mathcal{G}_{\text{gen}}$ (Eq.~\ref{gen_data_eq}), where $P$ structure these aesthetic information.
\begin{equation} 
\begin{aligned}
y &= \mathcal{G}_{\text{gen}}(x_v, P(s, \mu, \sigma, T))
\end{aligned}
\label{gen_data_eq}
\end{equation}
\begin{figure*}[t]
  \centering
  \includegraphics[width=\linewidth]{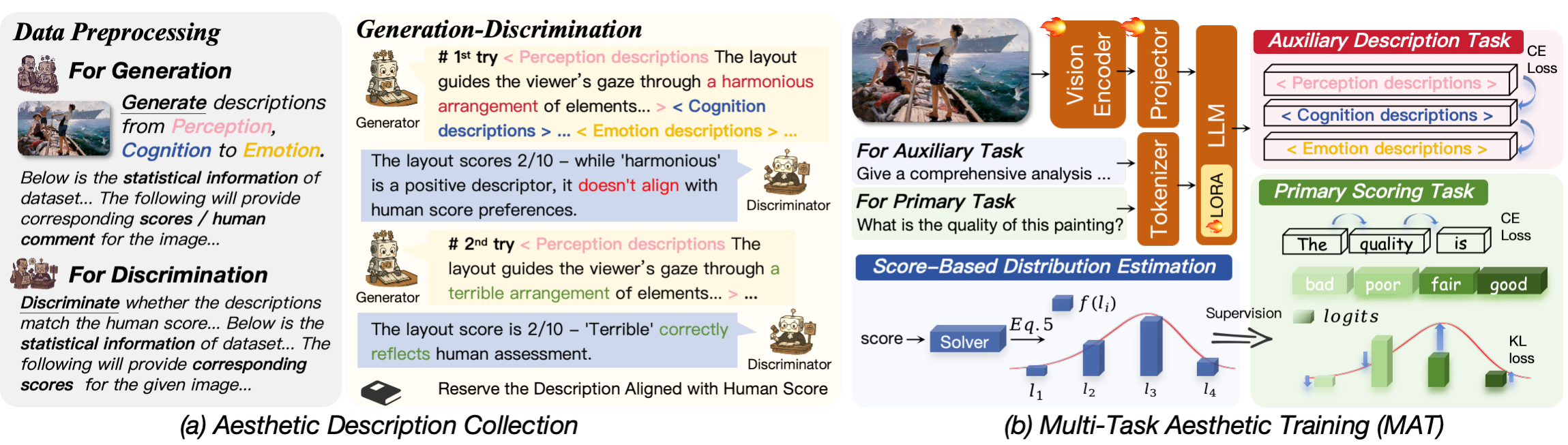}
  \caption{(a) We leverages a scalable, iterative framework to generate hierarchical aesthetic descriptions, ensuring the descriptions align with human scoring. (b) By employing Multi-Task Aesthetic Training to decompose hierarchical aesthetic elements and integrating a high-precision Score-Based Distribution Estimation method, ArtQuant achieves superior alignment with human artistic aesthetic scoring.}
  \label{train}
\end{figure*}
\noindent\textbf{Description discrimination.} Inspired by \textit{LLM-as-Judge~\cite{li2024generation}}, we propose an iterative framework to ensure the generated analyses are consistent with aesthetic scores (as shown in Fig.~\ref{train}a). The discriminator $\mathcal{G}_{\text{dis}}$ evaluates the consistency between the aesthetic scores and the generated analyses, thereby regulating output quality. Specifically, our generator $\mathcal{G}_{\text{gen}}$ is implemented using \textsc{GPT-4o}~\cite{yang2023dawn}, whose multimodal reasoning synthesizes analyses can effectively integrate visual feature analysis with aesthetic information; 
the discriminator $\mathcal{G}_{\text{dis}}$ employs DeepSeek-chat~\cite{bi2024deepseek} for alignment verification, ensuring that scores and visual content remain relatively independent. Although the current large language models inevitably have the problem of hallucination, we verified that our solution is still superior to human manual annotation in the experiment part. This is because human annotation is overly simplistic and generalized, whereas our method is more detailed and structured in a way that better aligns with the human aesthetic process.

\subsection{Aesthetic Acessment Learning}
\label{progressive}
\noindent\textbf{Auxiliary description task.}
To fully utilize the hierarchical analytical insights embedded in aesthetic descriptions, inspired by the inherent capabilities of MLLMs in generating textual outputs, we introduce a novel auxiliary task, aesthetic analysis generation, to systematically extract and integrate these insights into our framework. Specifically, we supervise our model to generate detailed aesthetic descriptions for images through supervised fine-tuning (SFT) on the RAD dataset's descriptions. Following standard practices in LLM training, we employ cross-entrophy loss (Eq.~\ref{ce_loss}) that predicts subsequent tokens for aesthetic descriptions generation.
\begin{equation}
\label{ce_loss}
    \mathcal{L}_{CE}=-\sum_{i = 1}^{T}\log P(t_i|t_1,t_2,\ldots,t_{i - 1}),
\end{equation}
where $t_i$ is the $i$-th token and $T$ is the sequence length.

\noindent\textbf{Primary scoring task.} For accurate score estimation using MLLMs, we convert the scores into token representations that align with the model’s output space~\cite{wu2023q,you2025teaching}. Let $i$ denote a discrete \emph{level} (e.g., excellent, good), $l_i$ be the \emph{score} associated with level $i$, and $p_i$ be the model's predicted probability for level $i$. The score $x$ is the expectation of the level scores, computed as follows:
\begin{equation}
\label{pred_score}
    x=\sum_{i}{p_{i}l_i}.
\end{equation}
To construct probability $p_i$ for ground truth $x_{gt}$, the probability distribution can be modeled as $f=N(\mu,\sigma)$ with $\mu$ as the Mean Opinion Score (MOS) and $\sigma$ as the variance of human annotations, where $d$ is the width of each level region~\cite{you2025teaching}:
\begin{equation}
    L(x) = \int_{l_{i}-\tfrac{d}{2}}^{l_{i}+\tfrac{d}{2}}f(x)dx.
\label{deqa}
\end{equation}
Unlike existing methods~\cite{you2025teaching,wu2023q} that rely on dataset variance information and risk introducing errors before training, we propose a \textbf{Score-Based Distribution Estimation} method. Our solution stems from a mapping perspective: during inference, the final score is computed as the expectation of discrete levels (Eq.~\ref{pred_score}), while training aims to align predicted level probabilities $p(i_{pred})$ with ground-truth distributions $p(i_{gt})$. 
Therefore, we can minimize the error between the expectation and the labels by optimizing the distribution parameters. However, due to the property that the sum of probabilities is 1, corresponding constraints need to be added during the solving process. The learning objective is formalized as:
\begin{equation}
\label{our_argmin}
    \begin{aligned}
        \mu^*, \sigma^* = \arg\min \left\| \sum f(l_i) l_i - x \right\|_2, \\
        \text{s.t.} \sum{f(l_i)}=1.
    \end{aligned}
\end{equation}
Here, we present the formulation where $f$ denotes the probability density function of the normal distribution:$f(x)=\frac{1}{\sigma\sqrt{2\pi}}e^{-\frac{(x - \mu)^{2}}{2\sigma^{2}}}$. We employ an automatic derivation of optimal mean and variance parameters through a numerical solver. This formulation enables our MLLM framework to model scores more effectively, improving prediction accuracy without sacrificing computational performance. 

For training, we adopt a hybrid loss combining cross-entropy (CE) and Kullback-Leibler (KL) divergence terms following DeQA~\cite{you2025teaching}. The CE loss $\mathcal{L}_{ce}$ supervises the prefix token sequence `\textit{The quality of the painting is}' to establish score-related text representations. In contrast, the KL loss (Eq.~\ref{l_kl}) minimizes the divergence between label constructed by Score-Based Distribution Estimation and predicted score distributions.
We combine these two loss terms into a unified objective, which we denote as the aesthetic score loss $L_{ASL}$ (Eq.~\ref{ASL}).
\begin{equation}
\label{l_kl}
    \mathcal{L}_{kl} = \sum_i p_i \log(\frac{p_i^{pred}}{p_i}),
\end{equation}
\begin{equation}
\label{ASL}
    \mathcal{L}_{ASL} = \mathcal{L}_{ce} + k\mathcal{L}_{kl}.
\end{equation}
\noindent\textbf{Training pipeline.}
In the description task, the vocabulary exhibits high diversity, far exceeding the predefined categories used for score regression. This may confuse the model when predicting the level tokens~\cite{you2025teaching}. To address this issue, we propose a progressive training strategy that effectively leverages the auxiliary description task to improve score prediction performance. Specifically, before training directly on the primary task, we design a multi-task aesthetic training (MAT) stage to capture the aesthetic characteristics of the artistic image. As formulated in Eq.~\ref{pt_loss}, this stage jointly optimizes both the auxiliary and primary objectives. By incorporating detailed aesthetic analysis, MAT enables the model to learn detailed artistic features more effectively, thereby providing a stronger initialization for artistic image aesthetics assessment.
\begin{equation}
\label{pt_loss}
    \mathcal{L}_{MAT} = \mathcal{L}_{CE} + k_{ASL}\mathcal{L}_{ASL}.
\end{equation}
To fully unlock the model’s scoring potential, in the second stage, we only train with $\mathcal{L}_ {ASL}$. This stage ensures the good representations learned in the MAT stage are stimulated and converge to the final score prediction task.

\section{Theoretical Analysis}
\label{sec:theory}
To explore why generating hierarchical descriptive text can improve the model's aesthetics assessment performance, we present the theoretical analysis underlying our auxiliary description learning framework. The analysis establishes formal guarantees for our approach while revealing fundamental relationships between visual representations, textual descriptions, and aesthetic scores.

\subsection{Formal Problem Setup}
Let us consider the artistic image assessment task through an information-theoretic lens. The input space $\mathcal{X}$ consists of visual artworks with their inherent features $x_v$, while the description space $\mathcal{D}$ contains the textual aesthetic analyses generated through our RAD pipeline. The target space $\mathcal{Y}$ represents discrete quality scores discretized into $K$ levels $\{l_1,...,l_K\}$. The latent space $\mathcal{Z}$ captures the multimodal representations learned by ArtQuant.

\subsection{Theoretical Guarantees}
Our first fundamental result establishes an upper bound on the conditional entropy of score prediction:

\begin{theorem}[Description-Score Dependency Bound]
\label{thm:dep-bound}
For any joint distribution $P(\mathcal{D},\mathcal{Y},\mathcal{Z})$ in our framework, the following inequality holds:
\begin{equation}
H(\mathcal{Y}|\mathcal{Z}) \leq H(\mathcal{D}|\mathcal{Z}) + H(\mathcal{Y}|\mathcal{D},\mathcal{Z}).
\end{equation}
\end{theorem}

This bound reveals that the uncertainty in predicting aesthetic scores ($H(\mathcal{Y}|\mathcal{Z})$) depends critically on two factors: the quality of description encoding ($H(\mathcal{D}|\mathcal{Z})$) and the conditional uncertainty of scores given descriptions ($H(\mathcal{Y}|\mathcal{D},\mathcal{Z})$). The ideal case occurs when textual descriptions $\mathcal{D}$ fully represent aesthetic scores $\mathcal{Y}$, leading to a simplified bound:

\begin{theorem}[Conditional Independence Bound]
\label{thm:indep-bound}
When $\mathcal{Y} \perp \mathcal{Z} | \mathcal{D}$ (conditional independence), we obtain:
\begin{equation}
H(\mathcal{Y}|\mathcal{Z}) \leq H(\mathcal{D}|\mathcal{Z}) + H(\mathcal{Y}|\mathcal{D}).
\end{equation}
\end{theorem}

In practical scenarios, perfect conditional independence may not hold. We therefore introduce an $\epsilon$-approximate independence condition (Theorem~\ref{thm:error-bound}) that relaxes the strict requirement while maintaining theoretical guarantees. This leads to our main error propagation result:

\begin{theorem}[Error Propagation Bound]
\label{thm:error-bound}
For $\epsilon = \epsilon_{\text{ver}} + \epsilon_{\text{gen}}$ is small enough, the prediction error satisfies:
\begin{equation}
\begin{aligned}
H(\mathcal{Y}|\mathcal{Z}) \leq & \underbrace{H(\mathcal{Y}|\mathcal{D}) + \epsilon\log|\mathcal{Y}| + H_2(\epsilon)}_{\text{description sufficiency}} + \\
& \underbrace{H(\mathcal{D}|\mathcal{Z})}_{\text{description generation ability}}.
\end{aligned}
\end{equation}
\end{theorem}

The implementation aspects that control these error terms correspond directly to key components of our framework. The generation error $\epsilon_{\text{gen}}$ can be ensured by the powerful GPT-4o for generation, while the verification error $\epsilon_{\text{ver}}$ is managed through score-comment alignment evaluation.

\subsection{Experimental Validation}
\begin{figure}[htbp]
    \centering
    \begin{subfigure}[b]{0.48\columnwidth}
        \includegraphics[width=\columnwidth]{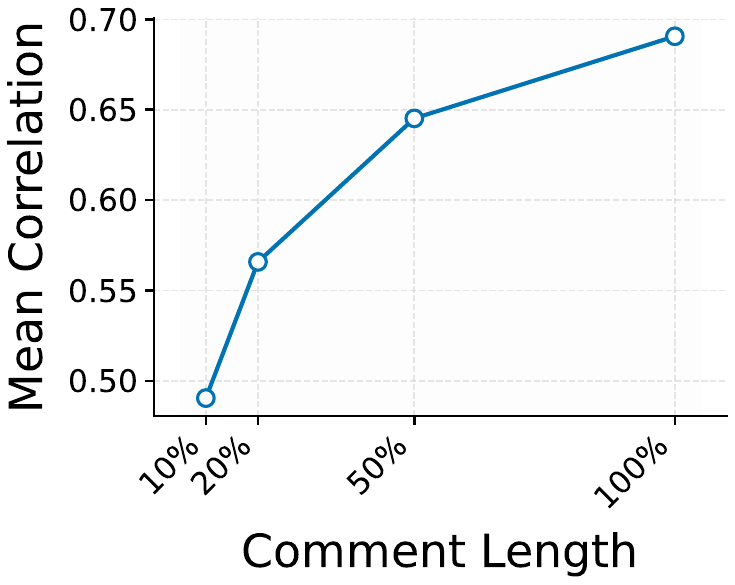}
        \label{fig:theory_exp_comment_length}
    \end{subfigure}
    \hfill
    \begin{subfigure}[b]{0.48\columnwidth}
        \includegraphics[width=\columnwidth]{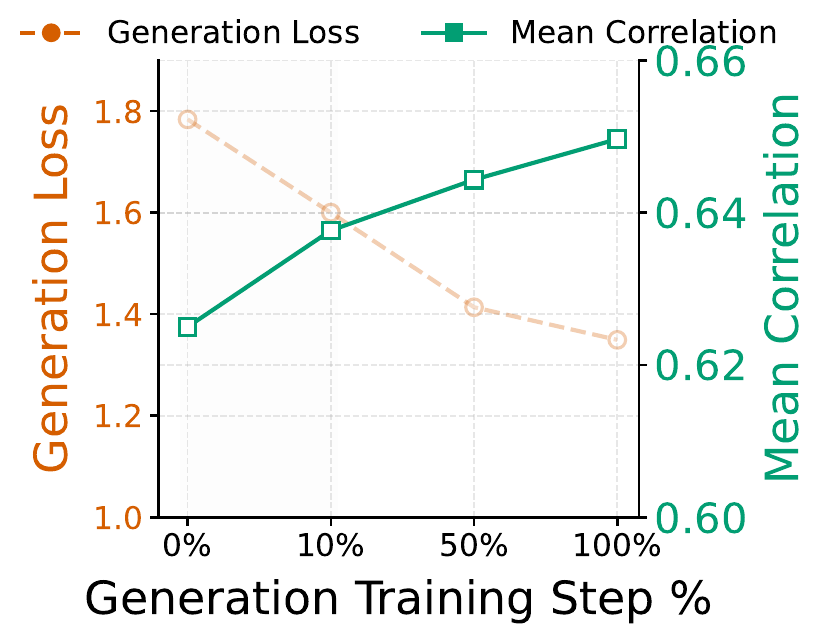}
        \label{fig:theory_exp_generation}
    \end{subfigure}
    \caption{
    Assessment performance with different description sufficiency and generation capability, where more descriptions and generation capability make the model's assessment more consistent with human.}
    \label{fig:theory_exp}
\end{figure}

\begin{table*}[t]
  \centering
  \begin{tabular}{@{}lcccccccl@{}}
    \toprule
    \multirow{2}{*}{Model} & 
    \multicolumn{2}{c}{\textbf{APDD}} & \multicolumn{2}{c}{\textbf{BAID}} & \multicolumn{2}{c}{\textbf{VAPS}} & 
    \multirow{2}{*}{Epochs} & \multirow{2}{*}{Modality} \\
    \cmidrule(lr){2-3} \cmidrule(lr){4-5} \cmidrule(lr){6-7}
    & SROCC$\uparrow$ & PLCC$\uparrow$ & SROCC$\uparrow$ & PLCC$\uparrow$ & SROCC$\uparrow$ & PLCC$\uparrow$ & & \\
    \midrule
    
    \multicolumn{9}{@{}l}{\textbf{Generalist MLLMs}} \\
    \hspace{0.5em}Qwen-VL-Plus~\cite{bai2023qwen} & 0.472 & 0.482 & 0.122 & 0.108 & -- & -- & -- & I+T+V \\
    \hspace{0.5em}Qwen-VL-Plus (+CoT) & 0.472 & 0.493 & 0.067 & 0.062 & 0.323 & 0.328 & -- & I+T+V \\
    \hspace{0.5em}GPT-4o~\cite{hurst2024gpt} & 0.531 & 0.558 & 0.122 & 0.108 & 0.170 & 0.197 & -- & I+T+A+V \\
    \hspace{0.5em}GPT-4o (+CoT) & 0.559 & 0.603 & 0.067 & 0.062 & 0.217 & 0.242 & -- & I+T+A+V \\
    \midrule
    
    \multicolumn{9}{@{}l}{\textbf{Specialist Models}} \\
    \hspace{0.5em}SAAN~\cite{yi2023towards} & 0.780 & 0.610 & 0.473 & 0.467 & 0.541 & 0.594 & 200 & I \\
    \hspace{0.5em}AANSPS~\cite{jin2024paintings} & 0.760 & 0.790 & -- & -- & -- & -- & -- & I \\
    \hspace{0.5em}ArtCLIP~\cite{jin2024apddv2} & 0.810 & 0.840 & -- & -- & -- & -- & 20+ & I+T \\
    \hspace{0.5em}LITA~\cite{sunada2024lita} & -- & -- & 0.490 & 0.573 & -- & -- & 15 & I+T \\
    \hspace{0.5em}EAMB-Net~\cite{10433197} & -- & -- & 0.496 & 0.487 & 0.567 & 0.628 & 100 & I \\
    \hspace{0.5em}AKA-Net~\cite{10810501} & -- & -- & 0.518 & 0.529 & 0.579 & 0.638 & 40 & I \\
    \hspace{0.5em}PVAFE~\cite{tang2025artistic} & -- & -- & 0.533 & 0.583 & -- & -- & 200 & I \\
    \midrule
    
    \multicolumn{9}{@{}l}{\textbf{Our Method}} \\
    \hspace{0.5em}ArtQuant & \textbf{0.871} & \textbf{0.894} & \textbf{0.543} & \textbf{0.589} & \textbf{0.625} & \textbf{0.681} & 4/2/8 & I+T \\
    \bottomrule
  \end{tabular}
  \caption{Score regression results across datasets (SROCC/PLCC metrics). Modality: I=Image, T=Text, A=Audio, V=Video.}
  \label{tab:performance_single}
\end{table*}
Our theoretical framework establishes that the model's artistic assessment capability (quantified by $H(\mathcal{Y}|\mathcal{Z})$) is fundamentally bounded by two factors: (1) the quality of generated descriptions ($H(\mathcal{D}|\mathcal{Z})$) and (2) the sufficiency of descriptions for assessment ($H(\mathcal{Y}|\mathcal{D})$). To validate these relationships, we conduct a series of controlled experiments. For the artistic image assessment task, model performance is quantified using the average of PLCC and SROCC metrics. The descriptive capability of the model is evaluated through the generation loss ($L_{CE}$), while description sufficiency is assessed by analyzing outputs of varying lengths. In the experiment on the relationship between the description ability and performance, we remove $L_{ASL}$ in the MAT stage to prevent interference caused by the participation of the score task training. All the results in the figure are the mean values of the performance on the APDD, BAID, and VAPS datasets.

Fig.~\ref{fig:theory_exp} (left) demonstrates the critical role of description sufficiency. As comment length increases from 10\% to 100\%, we observe a near-linear improvement in mean correlation scores (from 0.49 to 0.69, $r=0.92$). This strong positive relationship confirms our theoretical prediction that richer descriptions reduce       $H(\mathcal{Y}|\mathcal{D})$, thereby reducing the upper bound on assessment accuracy. 

Fig.~\ref{fig:theory_exp} (right) reveals the dual-axis relationship between description generation ability and model performance. As step size increases from 0 to 100\%: Generation loss (orange dashed line) decreases and performance (green solid line) increases from 0.62 to 0.65. The inverse correlation ($r=-0.99$) between generation loss and assessment performance empirically validates $H(\mathcal{D}|\mathcal{Z})$ as an upper bound for $H(\mathcal{Y}|\mathcal{Z})$. This demonstrates that improved description generation directly enables better artistic assessment.

\section{Experiments}
\subsection{Implementation Details}
\begingroup
\noindent\textbf{Dataset.}
To demonstrate the robustness of our method in predicting aesthetic scores, we employ three distinct datasets with significant variations. These datasets span professional expert annotations, crowd-sourced user preferences, and historical artwork assessments by non-specialized raters, collectively representing diverse aesthetic evaluation scenarios across temporal, cultural, and methodological dimensions. 

\noindent\textbf{Evaluation metrics.}
We evaluate AIAA performance using two metrics: Pearson’s correlation coefficient (PLCC) and Spearman’s rank correlation coefficient (SROCC) to measure prediction-ground truth alignment.

\noindent\textbf{Training setting.}
For the final model, in the MAT phase, the APDD, BAID, and VAPS are 3, 1, and 1 epoch(s) respectively, while in the SOT phase, the APDD, BAID, and VAPS are 1, 1, and 7 epoch(s) respectively. On all datasets, both $k$ and $k_{ASL}$ are 1. All experiments are conducted on four NVIDIA RTX 4090 GPUs, with total training times of 1 hour (APDD), 1.8 hour (BAID), and 20 minutes (VAPS). Notably, ArtQuant completes training in just 7.2 GPU hours on RTX 4090 for BAID, representing a substantial improvement over SAAN's~\cite{yi2023towards} 48-hour training time on RTX 3090 for the same dataset. 
\subsection{Quantitative Experiment}
\noindent\textbf{Art image quality assessment.}
In this section, we compare our model with two generalist MLLMs (GPT-4o~\cite{hurst2024gpt}, Qwen-VL-Plus~\cite{bai2023qwen}) and several specialist methods across multiple datasets. As shown in Table~\ref{tab:performance_single}, our model demonstrates superior performance. These experimental results demonstrate that ArtQuant, through pre-training on rich aesthetic comments, has learned fine-grained aesthetic representations, thereby enabling more accurate prediction of aesthetic scores. 
In terms of convergence efficiency, our method requires few training epochs to converge, significantly reducing computational costs compared to classic methods. This efficiency likely stems from the rich prior knowledge of LLM, particularly advantageous for smaller datasets like VAPS.
While GPT-4o~\cite {hurst2024gpt} and Qwen-VL-Plus~\cite{bai2023qwen} show that scale and inference strategies (e.g., chain-of-thought) contribute to performance gains on APDD and VAPS, their overall results remain weaker than both our method and classic approaches. This suggests that general MLLMs currently lack specialized aesthetic assessment capabilities. The prediction limitations of general MLLMs on BAID could potentially be attributed to the narrow score distribution of the dataset.

\subsection{Qualitative Experiment}
\begin{figure}[t!]
  \centering
  \includegraphics[width=\linewidth]{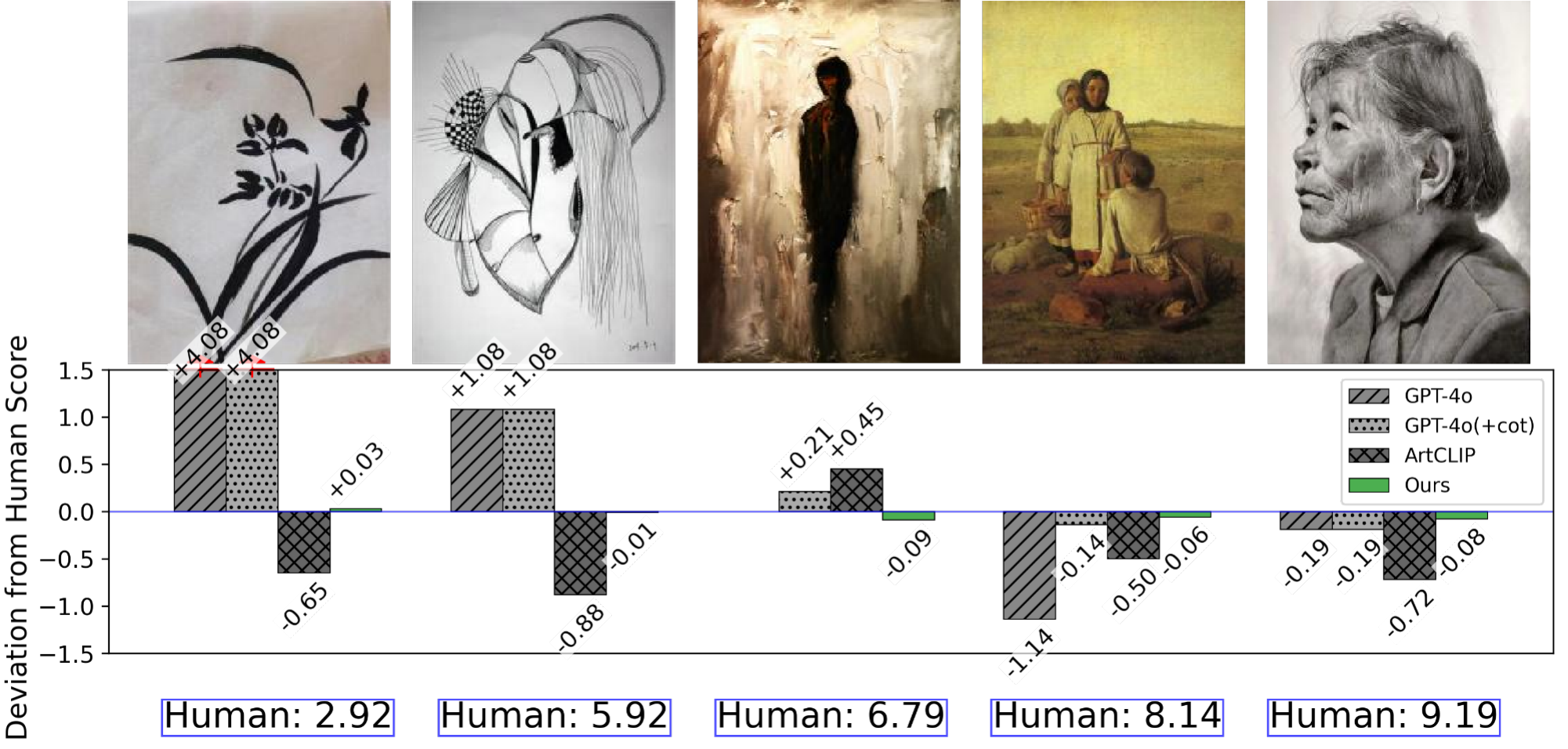}
  \caption{Qualitative results on the APDD test set. The vertical axis represents the error between model predictions and human scores, while the horizontal axis shows artistic images of varying aesthetic quality. Our method achieves better alignment with human judgments than other models across across paintings of different qualities.}
  \label{img_score}
\end{figure}
Our visual analysis (Fig.~\ref{img_score}) shows that ArtQuant achieves accurate predictions across varying aesthetic quality levels. Besides, generalist MLLMs tend to overestimate low-scoring artworks (evident in their 4.08 deviation for the lowest-scored painting), which may be related to the lack of fine-grained artistic discrimination. The divergence between methods diminishes for high-quality artworks, implying that exceptional aesthetic merit is more universally recognizable across different models.
\subsection{Ablation Experiments}
\label{Ablation}
\begin{table}[tbp]
  \centering
  \resizebox{\columnwidth}{!}{
      \begin{tabular}{l>{\quad}cc@{\qquad}cc}
        \toprule
        & \multicolumn{2}{c}{\textbf{SROCC}} & \multicolumn{2}{c}{\textbf{PLCC}} \\
        \cmidrule(lr){2-3}\cmidrule(lr){4-5}
        \textbf{Components} & Value & $\Delta_{\text{human}}$ & Value & $\Delta_{\text{human}}$ \\
        \midrule
        Human & 0.837 & \multicolumn{1}{c}{--} & 0.864 & \multicolumn{1}{c}{--} \\
        \addlinespace[0.5em]
        Perception & 0.865 & +3.35\% & 0.888 & +2.78\% \\
        + Cognition & 0.866 & +3.46\% & 0.890 & +3.01\% \\
        ++ Emotion & \textbf{0.871} & \textbf{+4.06\%} & \textbf{0.894} & \textbf{+3.47\%} \\
        \bottomrule
      \end{tabular}
  }
  \caption{Progressive hierarchical ablation results with $\Delta_{\text{human}}$ indicating percentage improvements relative to human baseline performance on APDD dataset.}
  \label{tab:percentage_ablation}
\end{table}

\begin{table}[t]
  \resizebox{\columnwidth}{!}{
  \begin{tabular}{lcccc}
    \toprule
    \textbf{Dataset} & \textbf{w/o MAT} & \textbf{w/ MAT} & \textbf{Gain (\%)} \\
    \midrule
    APDD & 0.863/0.889 & 0.871/0.894 & +0.9\%/+0.5\% \\
    BAID & 0.499/0.580 & 0.543/0.589 & +8.82\%/+1.55\% \\
    VAPS & 0.545/0.634 & 0.625/0.681 & +14.68\%/+7.41\% \\
    \bottomrule
  \end{tabular}
  } 
  \caption{Performance comparison and percentage improvements of whether train with MAT stage.}
  \label{tab:improvement_analysis}
\end{table}
\noindent\textbf{Annotation method.} To validate the benefits of LLM-assisted data generation for score prediction, we compared APDD manual annotations with the APDD subset of the RAD dataset under identical settings. As shown in Table~\ref{tab:percentage_ablation}, our method achieves significantly higher alignment with human judgments compared to simple and ambiguous manual annotations, with a 4.06\% improvement in SROCC (0.871 vs. 0.837) and 3.47\% gain in PLCC (0.894 vs. 0.864). The progressive performance improvement across perception (SROCC: +3.35\%), cognition (SROCC: +3.46\%), and emotion components (SROCC: +4.06\%) confirms that structured, multi-level annotations better capture the nuances of image aesthetics. The results suggest that LLM-generated descriptions provide superior supervision for learning aesthetic representations.

\noindent\textbf{Progressive learning.} To analyze the role of MAT, we conduct comprehensive evaluations across three distinct artwork. As shown in Table~\ref{tab:improvement_analysis}, the inclusion of the MAT training stage consistently improves performance across all datasets. Such enhancements indicate that establishing initial hierarchical, semantic aesthetic grounding via descriptions effectively bolsters the subsequent capability of score regression.

\subsection{Feature Analysis}
Fig.~\ref{feature} presents a comparative analysis of vision encoder attention heatmap between ArtCLIP~\cite{jin2024apddv2} and our proposed model. Our attention map shows significant improvements over ArtCLIP by precisely highlighting contour lines (e.g., hair strands) that match the `charm, smooth lines' description, demonstrating stroke-aware focus. The attention peaks align with actual shadow regions (nose bridge and neck), validating proper light-dark handling through physically-grounded lighting attention. Moreover, our model exhibits stronger emotional saliency with focused activation on expressive features (eyes and mouth), effectively capturing the annotated emotional atmospheres.
\begin{figure}[t]
  \centering
  \includegraphics[width=\linewidth]{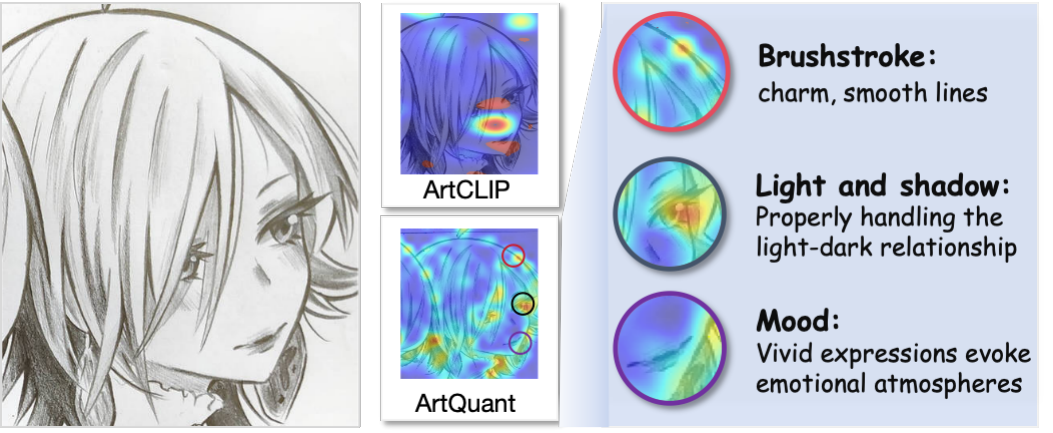}
  \caption{Comparative visualization of vision encoder heatmaps: original painting (left), ArtCLIP (top-center), and our method (bottom-center), with corresponding painting descriptions (right). Our method, enhanced by auxiliary description learning, achieves better semantic alignment in the heatmap representation.}
  \label{feature}
\end{figure}

\section{Conclusions}
We designs a scalable, iterative framework to generate hierarchical aesthetic descriptions that maintain strong alignment with human scoring. Based on the dataset, we utilizes Multi-Task Aesthetic Training to systematically decompose aesthetic elements across different levels, while incorporating high-precision Score-Based Distribution Estimation to ensure superior correlation with human artistic assessment. Extensive experiments demonstrate that ArtQuant achieves state-of-the-art performance across multiple benchmarks.



\section{Acknowledgments}
This work was supported in part by the National Natural Science Foundation of China under Grant U24B6012, 62406167, the Shenzhen Key Laboratory of Next Generation Interactive Media Innovative Technology, China (No. ZDSYS20210623092001004), and the National Science and Technology Council, Taiwan (No. 114-2221-E-006-114-MY3).

\bigskip

\bibliography{aaai}

\end{document}